\documentclass{article}

\usepackage{fontawesome}
\usepackage[final]{neurips_2024}
\usepackage{graphicx}
\usepackage[utf8]{inputenc} % allow utf-8 input
\usepackage[T1]{fontenc}    % use 8-bit T1 fonts
\usepackage{hyperref}       % hyperlinks
\usepackage{url}            % simple URL typesetting
\usepackage{booktabs}       % professional-quality tables
\usepackage{amsfonts}       % blackboard math symbols
\usepackage{nicefrac}       % compact symbols for 1/2, etc.
\usepackage{microtype}      % microtypography
\usepackage{xcolor}         % colors
\usepackage{multirow}

\title{Mind with Eyes: from Language Reasoning to Multimodal Reasoning}

\author{
Zhiyu Lin$^1$, \, 
Yifei Gao$^1$, \, 
Xian Zhao$^1$,\
Yunfan Yang$^1$, \
Jitao Sang$^1$\thanks{Corresponding Author}
\\
\vspace{0.1cm}
\{zyllin, yifegao, xianzhao, yunfanyang, jtsang\}@bjtu.edu.cn
\\
\vspace{0.1cm}
{$^1$Beijing Jiaotong University}
\\ 
{\vspace{0.1cm}
\faGithub: \small\url{https://github.com/ADaM-BJTU/Mind_with_eyes_Awesome_MLLMs_Reasoning}
}
}

\begin{document}

\maketitle

\begin{abstract}
Language models have recently advanced into the realm of reasoning, yet it is through multimodal reasoning that we can fully unlock the potential to achieve more comprehensive, human-like cognitive capabilities. This survey provides a systematic overview of the recent multimodal reasoning approaches, categorizing them into two levels: language-centric multimodal reasoning and collaborative multimodal reasoning. The former encompasses one-pass visual perception and active visual perception, where vision primarily serves a supporting role in language reasoning. The latter involves action generation and state update within reasoning process, enabling a more dynamic interaction between modalities. Furthermore, we analyze the technical evolution of these methods, discuss their inherent challenges, and introduce key benchmark tasks and evaluation metrics for assessing multimodal reasoning performance. Finally, we provide insights into future research directions from the following two perspectives: (i) from visual-language reasoning to omnimodal reasoning and (ii) from multimodal reasoning to multimodal agents. This survey aims to provide a structured overview that will inspire further advancements in multimodal reasoning research.
\end{abstract}

\section{Introduction}

Recently, Large Language Models (LLMs) have emerged as a cornerstone of language reasoning tasks~\cite{plaat2024reasoning,chen2025towards,zhang2024llm,bandyopadhyay2025thinking,wang2025tutorial,xu2025towards,chen2025towards}, leveraging their exceptional text generation and contextual comprehension capabilities. Models such as OpenAI-o1 \cite{gpto1} and Deepseek-R1 \cite{guo2025deepseek} have demonstrated human-like stepwise reasoning abilities in mathematical deduction, logical question answering, and code generation through strategies like Chain-of-Thought (CoT) prompting and reinforcement learning. However, the limitations of purely text-based reasoning are increasingly apparent: its input and output are confined to a single modality (text), rendering it inadequate for real-world scenarios requiring multimodal interactions (e.g., images, audio). With the advent of Multimodal Large Language Models (MLLMs)~\cite{bai2025qwen2,chen2024internvl,li2024llava,hu2024minicpm}, researchers have begun exploring the integration of LLMs' reasoning capabilities with visual, auditory, and other modalities. Early explorations \cite{zhang2023multimodal} attempted to transfer the CoT reasoning paradigm to vision-language tasks (e.g., visual question answering, chart parsing) via workflows that involve "parsing textual instructions, extracting image features, fusing multimodal representations, and generating reasoning conclusions."

In parallel, \emph{Knowledge Graph-based Multimodal Reasoning} has already established a systematic research framework~\cite{lee2024multimodal,zhu2022multimodal,zhang2022multimodal,zhu2022multi,chen2024knowledge}. By structuring knowledge representations (e.g., entity-relation-attribute triplets) and incorporating visual symbolic information, these methods explicitly model cross-modal semantic associations (e.g., linking image regions to textual descriptions) and enable interpretable reasoning through symbolic logic rules or graph neural networks (GNNs). Nevertheless, high-quality knowledge graphs heavily rely on costly human annotations and suffer from limited domain coverage, hindering their scalability to open-domain question answering and reasoning tasks. In contrast, MLLMs excel at learning implicit correlations from unstructured multimodal data, demonstrating superior generalization capabilities. Moreover, their powerful text generation abilities facilitate more natural human-AI interactions. Consequently, employing LLMs as the core engine for multimodal reasoning remains the prevailing paradigm.

\begin{figure*}
    \centering
    \includegraphics[width=1.0\linewidth]{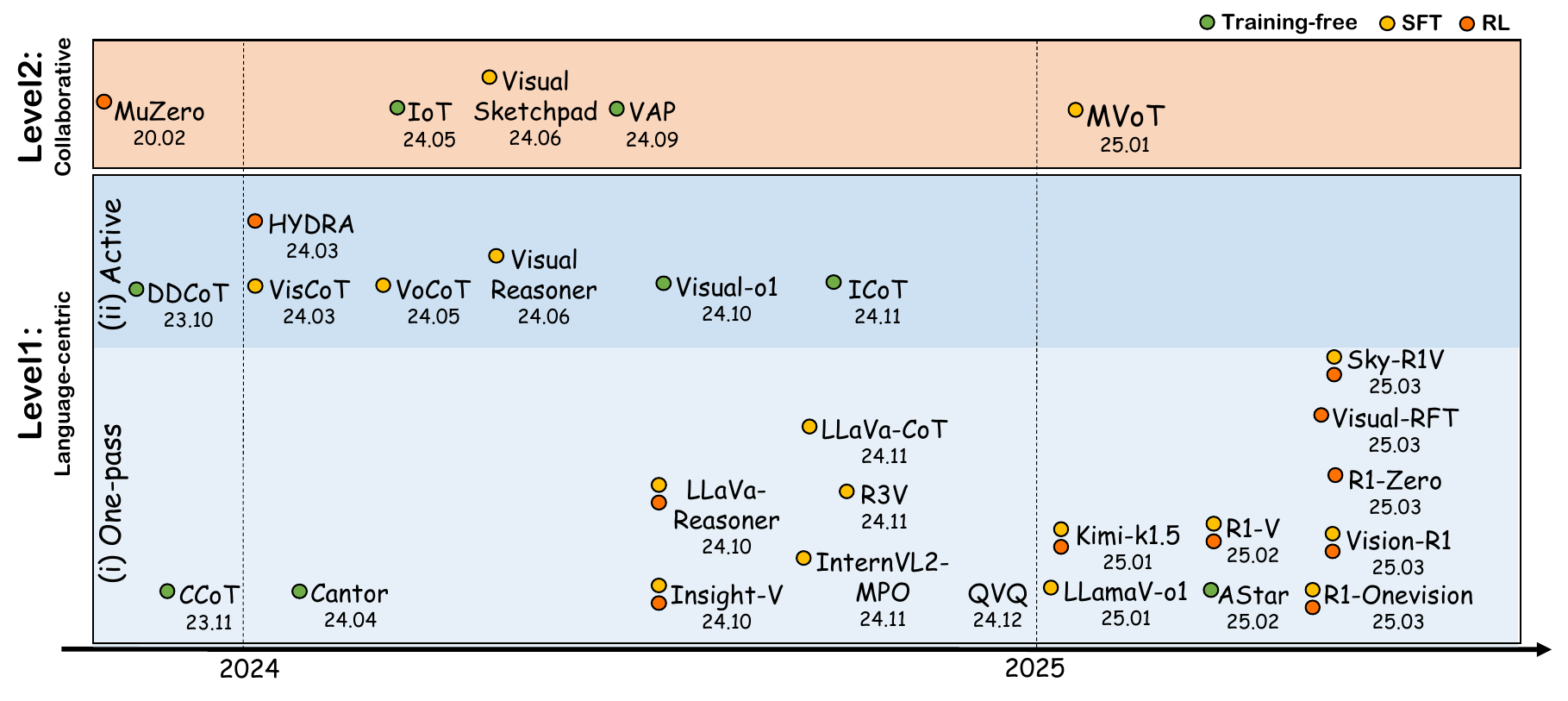}
    \caption{Timeline of multimodal reasoning models up to March 2025. We annotate and categorize the construction paradigms of reasoning models using dots of different colors.}
    \label{fig:fig1}
\end{figure*}

While LLM-based multimodal reasoning holds great potential, it faces more challenges than language-only reasoning. The core difficulties lie in ensuring cross-modal semantic alignment while enabling dynamic interaction and collaboration between modalities. First, vision and language differ significantly in information density and abstraction. Images convey detailed, low-level spatial features, whereas text captures high-level conceptual meaning. This disparity requires robust feature fusion mechanisms for cross-modal alignment. Second, the prevailing "vision-language feature concatenation + text generation" paradigm limits the generative potential of visual information and hinders dynamic cross-modal interaction. Effective multimodal reasoning requires models to adaptively prioritize the most relevant modality and guide information flow across modalities as needed. This survey systematically reviews recent multimodal reasoning methods and benchmark datasets as shown in Figure \ref{fig:fig1}. By examining technological progress and persistent challenges, we provide actionable insights to shape the next generation of multimodal reasoning systems.

\section{Taxonomy of Multimodal Reasoning}

\begin{figure*}
    \centering
    \includegraphics[width=0.92\linewidth]{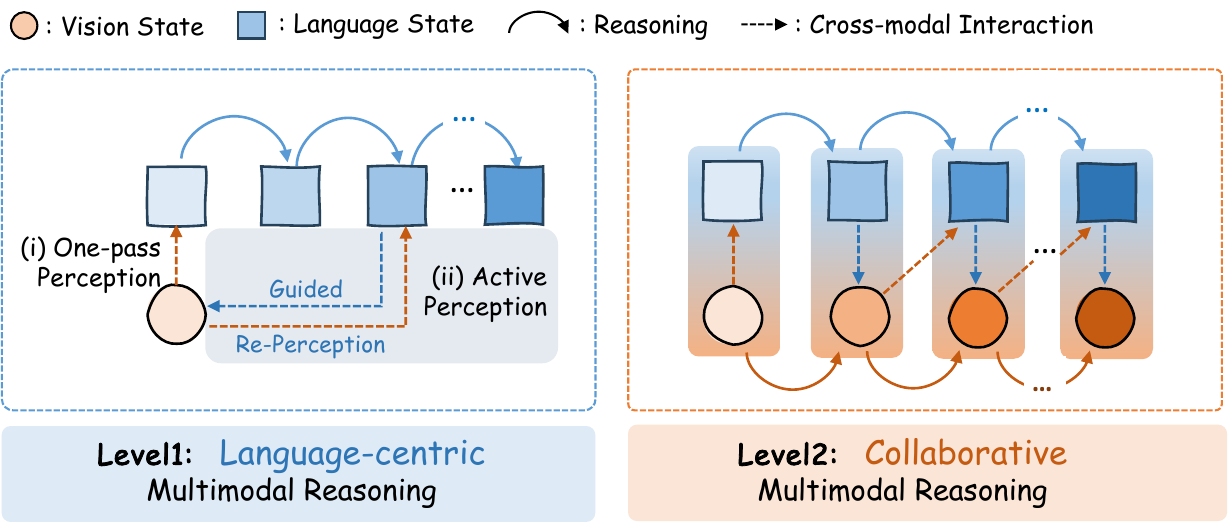}
    \caption{An analogous schematic diagram of two levels of multimodal reasoning.}
    \label{fig:fig2}
\end{figure*}

As previously discussed, multimodal reasoning is not an isolated activity within a single modality but rather a dynamic interaction process where vision and language modalities collaboratively deepen cognitive reasoning. Through systematic analysis of existing works, we observe that nearly all solutions explicitly construct reasoning chains within the language space, with their critical distinctions lying in the strategic hierarchy of visual information processing. Based on how visual modaility involves in the reasoning process, We categorize these approaches into two progressive levels as shown in Figure \ref{fig:fig2}

\subsection{Level 1: Language-centric Multimodal Reasoning}
In this paradigm, the visual modality primarily serves perceptual and feature extraction roles, while reasoning is entirely dominated and driven by the language modality. Based on the triggering mechanisms of visual perception, this level is divided into two subcategories:

\textbf{One-pass Visual Perception:} These methods treat visual information as static context. The model encodes images only once during the input stage ("only look once", e.g., CLIP-based global feature extraction \cite{xu2411llava}), relying solely on the language modality for subsequent information integration and logical deduction.

\textbf{Active Visual Perception:} Intermediate reasoning steps generated by the language modality trigger multiple rounds of visual re-perception (e.g., dynamic region cropping or zooming \cite{cheng2024least}). The model actively retrieves required visual details based on textual reasoning cues, forming a "look-back" mechanism. Compared to one-pass perception, this approach is proved to enhance the reliability of multimodal alignment during reasoning.

\subsection{Level 2: Collaborative Multimodal Reasoning}
When reasoning involves visual action reasoning and visual state updating, the visual modality transcends its passive perception role to engage in collaborative reasoning with the language modality. Key characteristics include:

\textit{Visual action reasoning:} The visual modality not only responds to language instructions but also autonomously generates internal reasoning actions (e.g., invoking visual tools for image editing or leveraging generative capabilities to reconstruct images \cite{zhou2024image}). This is marked by explicit reasoning trajectories within the visual feature space.

\textit{Visual state update:} By executing the above actions, the model dynamically updates visual contextual information (e.g., generating new geometric diagrams with auxiliary lines \cite{hu2024visual}). The updated visual representations feedback as new constraints to the language modality, triggering subsequent intermediate reasoning steps.

This taxonomy reveals the evolution of multimodal reasoning technologies: from language-dominated unilateral control toward vision-language co-reasoning. Below, we detail existing solutions and derive actionable insights for each level of works.

\section{One-pass Visual Perception}
\label{section2}

\subsection{Prompt-based Solutions}
This type of works \cite{gao2024cantor,mitra2024compositional,wu2025boosting}  initially configure the multimodal large model to assume roles as different functional modules through system messages. This is followed by generating intermediate reasoning results or strategies in a workflow manner, which are then used to synthesize the final answer. Depending on the definition of the functional modules, the form of the intermediate reasoning results can vary widely. We will illustrate this using the following three works as representatives.

\textbf{Cantor} \cite{gao2024cantor} deconstructs the visual reasoning task into two steps: decision generation and execution. In the first step, the multimodal large model is prompted to assume multiple roles, completing principle analysis, module selection, and task allocation. This enables the model to identify the necessary visual contextual information for different tasks through decomposition and analysis. In the second phase, the model is required to complete various subtasks, invoking the MLLM to generate corresponding high-level visual features based on task analysis. Finally, the results of the subtasks are synthesized and summarized to provide the final answer. The framework proposed in this article is characterized by dividing visual perception into two stages during the reasoning process: (1) analysis of visual contextual information requirements and (2) generation of high-level visual features.

\textbf{CCoT} \cite{mitra2024compositional} proposes using scene graphs to formally describe the results of visual reasoning. The advantage of scene graphs lies in their ability to provide a highly structured representation of visual objects, relationships, and attributes within an image, thus overcoming the limitations of purely textual descriptions. To address the high data collection cost associated with scene graphs, the article introduces a two-stage workflow to construct a compositional reasoning chain. Initially, MLLMs generate relevant scene graphs based on the image, question description, and scene graph requirements. Subsequently, the MLLM receives prompts involving the scene graph, image, and task to produce the final answer. The article emphasizes the importance of the relationships between objects and attributes in an image for solving visual reasoning tasks, highlighting that the proposed scene graph representation enhances the MLLM's reasoning about object and attribute relationships compared to captions.

\textbf{AStar} \cite{wu2025boosting}'s approach involves constructing "thought cards" as external explicit reasoning guidelines. Inspired by human behavior, AStar first defines six types of reasoning actions as the building blocks of thought cards. It then leverages the advantages of Monte Carlo Tree Search (MCTS) to sample a large amount of reasoning path data from a small set of raw data. By categorizing the reasoning actions, it labels all reasoning data to create the final thought cards. During the reasoning phase, heuristic prompts are constructed by retrieving the thought card most similar to the target problem, guiding MLLMs to output the reasoning steps and results for the target problem. The advantage of thought cards lies in the ability to generate a large and diverse set of reasoning path data through MCTS sampling. Compared to training methods that use sampled data, AStar achieves a compelling balance between performance and efficiency.

\begin{table*}[]
\centering
\caption{Summary of multimodal reasoning methods. \textbf{LCMR} stands for Language-centric Multimodal Reasoning, and \textbf{CMR} stands for Collaborative Multimodal Reasoning. “One-Pass” and “Active” refer to two types of visual perception methods.}
\vspace{2mm}
    \resizebox{1.0\linewidth}{!}{
\begin{tabular}{@{}c|c|lcccc@{}}
\toprule
\multicolumn{2}{c}{\textbf{Category}} & \textbf{Method} & \textbf{Year} & \textbf{Strategy} & \textbf{Task} & \textbf{Characteristic} \\ \midrule
\multicolumn{1}{c|}{\multirow{23}{*}{\begin{tabular}[c]{@{}c@{}}\textbf{LCMR}\end{tabular}}} & \multicolumn{1}{c|}{\multirow{15}{*}{\begin{tabular}[c]{@{}c@{}}\textbf{One-} \\ \textbf{Pass}\end{tabular}}} & CCoT \cite{mitra2024compositional} & 23.11 & Training-Free & General Reasoning & Reasoning with Scene Graph \\
\multicolumn{1}{c|}{} & \multicolumn{1}{c|}{} & Cantor \cite{gao2024cantor} & 24.04 & Training-Free & Science VQA, Math & High-level Information Cognition \\
\multicolumn{1}{c|}{} & \multicolumn{1}{c|}{} & Astar \cite{wu2025boosting} & 25.02 & Training-Free & General Reasoning & MCTS-Powered Thought Card \\
\multicolumn{1}{c|}{} & \multicolumn{1}{c|}{} & LlamaV-o1 \cite{thawakar2025llamav} & 25.01 & SFT & General Reasoning & VRC-Benchmark and Curriculum Learning \\
\multicolumn{1}{c|}{} & \multicolumn{1}{c|}{} & LLaVA-CoT \cite{xu2411llava} & 24.11 & SFT & General Reasoning & Stage-level Beam Search \\
\multicolumn{1}{c|}{} & \multicolumn{1}{c|}{} & LLaVA-Reasoner \cite{zhang2024improve} & 24.10 & SFT+RL & General Reasoning & 193K ShareGPT-4o-Reasoning Dataset \\
\multicolumn{1}{c|}{} & \multicolumn{1}{c|}{} & Insight-V \cite{dong2024insight} & 24.10 & SFT+RL & General Reasoning & Multi-agent and Iterative DPO \\
\multicolumn{1}{c|}{} & \multicolumn{1}{c|}{} & InternVL2-MPO\cite{wang2024enhancing}& 24.11 & SFT+RL & General Reasoning & Mixed Preference Optimization Loss \\
\multicolumn{1}{c|}{} & \multicolumn{1}{c|}{} & R1-V \cite{chen2025vinci} & 25.02 & SFT+RL & Visual Counting & Reasoning Data from DeepSeek R1 \\
\multicolumn{1}{c|}{} & \multicolumn{1}{c|}{} & R1-Onevision \cite{yang2025r1onevision} & 25.03 & SFT+RL & General Reasoning & 155k CoT Data from Deepseek-R1 \\
\multicolumn{1}{c|}{} & \multicolumn{1}{c|}{} & Mulberry \cite{yao2024mulberry} & 24.12 & SFT & General Reasoning & Collective MCTS and Self-training \\
\multicolumn{1}{c|}{} & \multicolumn{1}{c|}{} & R3V \cite{cheng2024vision} & 24.11 & SFT & General Reasoning, Web Navigation & Iterative SFT with Self-refine Loss \\
\multicolumn{1}{c|}{} & \multicolumn{1}{c|}{} & Vision-R1 \cite{huang2025vision} & 25.03 & SFT+RL & General Reasoning& 200K CoT Data and PTST Strategy \\
\multicolumn{1}{c|}{} & \multicolumn{1}{c|}{} & R1-Zero \cite{zhou2025r1} & 25.03 & RL & General Reasoning & Applying RL on 2B non-SFT MLLM \\
\multicolumn{1}{c|}{} & \multicolumn{1}{c|}{} & Visual-RFT \cite{liu2025visual} & 25.03 & RL & Object Detection, Visual Grounding, Classification & Visual RFT on Limited Data \\
\multicolumn{1}{c|}{} & \multicolumn{1}{c|}{} & QVQ \cite{qvq-72b-preview} & 24.12 & -- & General Reasoning & --\\
\multicolumn{1}{c|}{} & \multicolumn{1}{c|}{} & Kimi k1.5 \cite{team2025kimi} & 25.01 & SFT+RL & General Reasoning, Code Generation & Long Context Scaling during RL \\
\multicolumn{1}{c|}{} & \multicolumn{1}{c|}{} & Skywork-R1V \cite{skywork2025r1v} & 25.03 & SFT+RL & General Reasoning & Iterative SFT and GRPO \\ \cmidrule(l){2-7} 
\multicolumn{1}{c|}{} & \multicolumn{1}{c|}{\multirow{8}{*}{\begin{tabular}[c]{@{}c@{}}\textbf{Active}\end{tabular}}} & DDCoT \cite{zheng2023ddcot} & 23.10 & Training-Free & Science VQA & Visual perception with VQA models \\
\multicolumn{1}{c|}{} & \multicolumn{1}{c|}{} & HYDRA \cite{ke2024hydra} & 24.03 & RL & General Reasoning, Visual Grounding & Visual perception with foundation models \\
\multicolumn{1}{c|}{} & \multicolumn{1}{c|}{} & FAST \cite{sun2024visual} & 24.08 & SFT & General Reasoning, Referencing Segmentation & Fast and slow thinking mode \\
\multicolumn{1}{c|}{} & \multicolumn{1}{c|}{} & VisualReasoner \cite{cheng2024least} & 24.06 & SFT & General Reasoning & Bottom-up training data synthesis \\
\multicolumn{1}{c|}{} & \multicolumn{1}{c|}{} & ICoT \cite{gao2024interleaved} & 24.11 & Training-Free & General Reasoning & CoT with relevant visual tokens \\
\multicolumn{1}{c|}{} & \multicolumn{1}{c|}{} & Visual-o1 \cite{ni2024visual} & 24.10 & Training-Free & General Reasoning, Referencing Segmentation & Multi-turn reasoning \\
\multicolumn{1}{c|}{} & \multicolumn{1}{c|}{} & VoCoT \cite{li2024vocot} & 24.05 & SFT & General Reasoning, Spatial Reasoning & Key object-centered CoT \\
\multicolumn{1}{c|}{} & \multicolumn{1}{c|}{} & VisCoT \cite{shao2024visual} & 24.03 & SFT & General Reasoning, Visual Grounding & CoT with key regions \\ \midrule
\multicolumn{2}{c|}{\multirow{5}{*}{\textbf{CMR}}} & MVoT \cite{li2025imagine} & 25.01 & SFT & Spatial Reasoning & State updates by model itself \\
\multicolumn{2}{c|}{} & MuZero \cite{schrittwieser2020mastering} & 20.02 & RL & Go, Atari & State update and action generation together \\
\multicolumn{2}{c|}{} & VAP \cite{xiao2024enhancing} & 24.09 & Training-Free & General Reasoning & State update by tools \\
\multicolumn{2}{c|}{} & Visual Sketchpad \cite{hu2024visual} & 24.06 & Training-Free & General Reasoning & State update by tools \\
\multicolumn{2}{c|}{} & IoT \cite{zhou2024image} & 24.05 & Training-Free & General Reasoning & State update by tools \\ \bottomrule
\end{tabular}
}
\end{table*}

\subsection{Learning-based Solutions}

This type of works~\cite{thawakar2025llamav,xu2411llava,zhang2024improve,shao2024visual,dong2024insight,wang2024enhancing,chen2025vinci,yang2025r1onevision,yao2024mulberry,cheng2024vision,huang2025vision,zhou2025r1,liu2025visual} involve first constructing a multimodal training dataset that includes reasoning chains, followed by applying Supervised Fine-Tuning and reinforcement learning to MLLMs. We will elaborate on the data construction methods and training paradigms.

\subsubsection{Data Construction}

A common approach involves utilizing powerful teacher models (e.g., GPT-4o, Deepseek-R1) for knowledge distillation and data filtering, including chain-of-thought reasoning and scoring of reasoning results or each reasoning step. Recent works~\cite{thawakar2025llamav,xu2411llava,zhang2024improve,shao2024visual} employs a "Let's think step by step" prompt to input VQA data into GPT-4o, obtaining chain-of-thought outputs and answers. Additionally, \cite{dong2024insight,wang2024enhancing} select robust open-source models such as QwenVL and InternVL for data filtering. Recent work has also explored using advanced language reasoning models for knowledge distillation. To address the challenge of language models being unable to process image modalities, \cite{chen2025vinci} convert images into captions, while \cite{yang2025r1onevision} design a formalized text grammar to describe musical scores, tables, and images. Once images are expressed in text, they are input along with the original questions into language reasoning models to obtain solutions in a chain-of-thought format. The advantage of this method is leveraging the strong reasoning capabilities of language models, while the challenge lies in the potential information loss when converting images to text. Another series of works~\cite{yao2024mulberry, cheng2024vision} involves the policy model (MLLMs) autonomously sampling and generating reasoning path samples, followed by iterative self-training. \cite{yao2024mulberry} proposes implementing MCTS within a policy model ensemble composed of multiple MLLMs to enhance the diversity of reasoning path data, constructing reflective paths based on negative nodes in the sampling process to supplement training data. \cite{cheng2024vision} similarly utilize the numerous negative samples generated during sampling to create reflective training data, thereby guiding the enhancement of the model's reflective capabilities.

\subsubsection{Training Paradigm}

\textit{SFT Only}~\cite{xu2411llava,yao2024mulberry,thawakar2025llamav,shao2024visual,cheng2024vision}. This approach involves training MLLMs using SFT, with chain-of-thought reasoning and answers as prediction targets. The models are fine-tuned using a Next Token Prediction loss function. Futhermore, the concept of curriculum learning is employed by \cite{thawakar2025llamav}, implementing a two-stage SFT training method for MLLMs, progressing from easy to difficult tasks: initially completing simple image captioning tasks, followed by complex multimodal reasoning tasks. \cite{yao2024mulberry,cheng2024vision} propose iterative self-training using reasoning path samples obtained by the model itself, enhancing the quality of sampled data and reasoning capabilities through continuous reflection.

\textit{SFT + RL-based Training}~\cite{zhang2024improve,dong2024insight,yang2025r1onevision,chen2025vinci,wang2024enhancing,huang2025vision}. The training process is divided into two phases. In the first phase, the aforementioned SFT training is conducted on MLLMs to obtain policy model $\pi_{\theta}$. In the second phase, a preference dataset is sampled based on $\pi_{\theta}$, and DPO~\cite{zhang2024improve,dong2024insight,wang2024enhancing} or GRPO~\cite{yang2025r1onevision,chen2025vinci,huang2025vision} methods are typically used to obtain the optimized policy model $\pi_{\theta}^*$. Furthermore, an iterative DPO training method is employed by \cite{dong2024insight}, executing multiple rounds of data sampling and DPO to better simulate online reinforcement learning. To improve training efficiency, \cite{wang2024enhancing} propose the MPO training framework, which combines Next Token Prediction and DPO loss functions in a single training process to achieve both SFT and reinforcement learning objectives simultaneously.

\textit{RL-based Training Only}\cite{liu2025visual,zhou2025r1}. GRPO is directly applied to non-SFT models. Visual RFT \cite{liu2025visual} designed three verifiable reward functions related to visual tasks: IOU and confidence rewards for object detection tasks, and classification accuracy rewards for classification tasks. This work found that combining reinforcement fine-tuning with policy models enhances few-shot learning capabilities and generalization, offering significant advantages over SFT in data-limited scenarios. R1-zero \cite{zhou2025r1} validate the feasibility of this training strategy on small parameter models (Qwen2-VL-2B) and replicated the "aha moment" during experiments.

\section{Active Visual Perception}

multimodal reasoning tasks depend on extracting rich and multi-grained visual information from images. However, current Multimodal Large Language Models (MLLMs) exhibit limited visual comprehension capabilities, making one-pass visual perception—examining the image only once during the problem formulation stage—potentially insufficient for acquiring the critical visual information required for reasoning. Specifically, 1) one-pass visual perception is difficult to realize fine-grained visual semantic understanding, leading to hallucinations and misunderstandings during reasoning~\cite{chen2022visualgpt,zhou2025they,huang2024opera,chen2023mitigating}; 2) one-pass visual perception struggles to comprehensively capture multi-granularity and multi-region visual cues in the image, resulting in the omission of key information~\cite{yin2023lamm,zhao2024mg,chen2024lion}. To overcome the limitations of single visual perception, some studies~\cite{gao2024interleaved,ni2024visual,li2024vocot,zheng2023ddcot,ke2024hydra,sun2024visual,cheng2024least} incorporate multiple rounds of visual perception into the multimodal reasoning process. Just as humans need to examine images multiple times to fully understand them, these methods involve iteratively extracting various levels of visual information from images throughout the reasoning process, thereby achieving enhanced performance on reasoning task.

\subsection{Self Iterative Perception}

This type of works utilizes the intrinsic visual capabilities of MLLM to perform multiple rounds of visual perception. ICoT \cite{gao2024interleaved} generates reasoning steps with paired visual and textual rationales to express the fine-grained associations between the image and the reasoning process. Specifically, it selects top-k relevant visual tokens based on attention maps and inserts them into the reasoning steps, making it adaptable to various MLLMs without additional training. Visual-O1 \cite{ni2024visual} employs multi-turn chain-of-thought reasoning to assist models in correctly understanding ambiguous instructions. During each iteration, the model leverages previous disambiguation experience to analyze the image and instruction, and reflects on the reasoning process to update the disambiguation experience. This iterative cycle continues until the model achieves a clear understanding of the instruction, allowing it to output the correct answer. In addition, the scarcity of multi-step reasoning data with interleaved-modal reasoning steps is an important issue that hinders the training of MLLMs for multimodal reasoning. To address this, some works construct chain-of-thought (CoT) datasets that incorporate multiple visual perceptions through a well-designed generation process. VisCoT \cite{shao2024visual} creates a CoT dataset annotated with bounding boxes highlighting key regions essential for answering the questions. During the reasoning process, the MLLM initially predicts key regions with bounding boxes and incorporates the visual features of these regions.  Similarly, VoCoT \cite{li2024vocot} constructs a dataset with key object-centered chains-of-thoughts, where key objects are represented in the format of “<text description, coordinates, visual object representation>” in each reasoning step. By performing supervised fine-tuning (SFT) on these datasets, the models generate multimodal chains of thought in the specified format, simultaneously improving the model's visual perception and reasoning capabilities. 

\subsection{Tool-assisted Iterative Perception}

Given the limitations in visual processing capabilities of MLLMs, this type of works employs specialized perception tools to perform multiple rounds of visual perception. Early multimodal reasoning methods were based on LLMs. Since LLMs are unable to accept image as input, such methods usually convert images into captions as visual input and use specialized visual models to achieve further visual perception in the reasoning process. DDCoT \cite{zheng2023ddcot} decomposes the question into sub-questions based on LLM and answers the uncertain sub-questions using a VQA model. The LLM then integrates the information and generates a rationale with critical thinking, resulting in more accurate and interpretable multimodal reasoning. HYDRA \cite{ke2024hydra} utilizes a LLM planner to generate multiple instruction samples, each with varying complexity and validity probabilities. An RL agent then evaluates and selects the most promising samples for execution. Next, the LLM reasoner translates these instructions into executable Python code, which leverages vision foundation models (e.g., GLIP, BLIP2, XVLM) to perform visual tasks such as object detection, verification, and captioning.  The resulting visual information is converted to text format and stored in the Memory Bank to support incremental reasoning

Additionally, some MLLM-based methods integrate task-specific models for more accurate visual perception. FAST \cite{sun2024visual} employs a switch adapter to dynamically select between fast and slow thinking mode, tailoring the problem-solving approach to different task complexity. For the slow thinking mode, two specific adapters are trained to generate target regions and pixel-level segmentations to compose multimodal chains of evidence. VisualReasoner \cite{cheng2024least} adopts a bottom-up approach to automatically generate questions and multi-step reasoning paths for images, followed by supervised fine-tuning on this synthetic CoT dataset. It divides the complex synthesis task into a few simple sub-tasks, and relies on open-sourced models to accomplish the sub-tasks, such as grounding, highlighting and Optical Character Recognition (ORC).

\textbf{Discussions.} The existing methods introduce multiple rounds of visual perceptions based on the intrinsic visual capabilities of MLLM or by invoking specialized visual models to executing sub-tasks like grounding, highlighting, and OCR. By performing visual perception multiple times during reasoning, the model effectively extracts rich and multi-grained information from the image, thereby enhancing its multimodal reasoning capability and improving interpretability. However, the effectiveness of these methods is still constrained by the visual capabilities of MLLMs and specialized visual models. Besides, they face challenges with tasks that require visual state updating during the reasoning process, such as spatial reasoning.

\section{Collaborative Multimodal Reasoning}

While large language models demonstrate strong reasoning capabilities, they typically treat visual information as static context rather than actively integrating it into the reasoning process. In this section, we summarize a series of approaches~\cite{li2025imagine,schrittwieser2020mastering,xiao2024enhancing,lin2025investigating,hu2024visual,zhou2024image} where visual information is modified during the reasoning process, a concept we refer to as collaborative multimodal reasoning. The essence of collaborative multimodal reasoning lies in simultaneously generating actions in the language space and updating states in the visual space. Similar to human cognitive processes, humans flexibly integrate both language and visual reasoning when solving problems. For example, in the game of Go, a player must not only decide on the next move (action generation) but also simultaneously understand and simulate the evolving board state after each move (state update). Based on this definition, the implementation of collaborative multimodal reasoning can be generally categorized into the following two approaches.

\subsection{State Update without Action Reasoning}

In this scenario, the model's reasoning process is reflected in predicting the next-state representation. This requires a multimodal architecture capable of visual generation. However, due to the limitations of current models' visual generation capabilities, existing research primarily focuses on simple tasks. In the spatial reasoning task, MVoT \cite{li2025imagine} overcomes the limitations of traditional language reasoning by incorporating visual state changes into the reasoning process. By enhancing its image generation capabilities, the model can visualize its reasoning process, enabling it to "think" in a more unified and coherent manner, rather than relying solely on language space.

\subsection{Joint Action Reasoning and State Update}

In this scenario, the model first generates actions to accomplish a well-defined reasoning objective and then updates the state based on the generated actions. In early research, MuZero \cite{schrittwieser2020mastering} facilitates state transitions and action generation by dynamic and prediction functions. The action generation and state updates are performed directly on the latent space, allowing it to operate without direct feedback from the real environment. Realizing the limitations of visual generative capabilities in complex and sequential reasoning, recent studies have attempted to leverage external tools to update visual states. Specifically, models are equipped with a visualization canvas and tools for drawing, allowing them to refine visual representations in the reasoning process. In mathematical reasoning tasks, libraries such as Matplotlib and Seaborn~\cite{xiao2024enhancing,lin2025investigating,hu2024visual} are employed to generate plots or add auxiliary lines, while in visual tasks, segmentation models are used to generate masks \cite{zhou2024image}. Without the need for additional post-training of multimodal language models, these modified visual representations significantly improve the model’s reasoning capability cooperating with carefully designed prompt, enabling more effective solutions to multimodal tasks.

\textbf{Discussions}: The reasoning ability demonstrated by the language model is based on the powerful generation ability of the model itself. However, the deficiency of multimodal models in visual generation has led to the fact that collaborative multimodal reasoning has not made sufficient progress. It should be pointed out that the process of state update through external tools is not the final form of multimodal reasoning. Future research should first focus on designing a model architecture suitable for multimodal reasoning, integrating visual and language generation to achieve a more seamless and autonomous reasoning process.

\section{Benchmark}

\begin{table}[htbp]

  \begin{center}

    \caption{Summary of multimodal reasoning benchmarks. \textbf{General} stands for general reasoning task, \textbf{Academic} stands for academic-based task, \textbf{Spatial} stand for spatial reasoning task, and \textbf{Logical} stands for logical reasoning task.}
    \label{tab:tab2}
    \renewcommand{\arraystretch}{1.3}
    \resizebox{0.9\columnwidth}{!}{
    \begin{tabular}{c|c|cccc|c|c} % <-- Alignments: 1st column left, 2nd middle and 3rd right, with vertical lines in between
    \toprule
      \multirow{2}{*}{\textbf{Benchmark}} & \multirow{2}{*}{\textbf{Year}} & \multicolumn{4}{c|}{\textbf{Task}}& \multirow{2}{*}{\textbf{Metrics}}& \multirow{2}{*}{\textbf{Samples}}\\
      \cline{3-6}
      & & \textbf{General} & \textbf{Academic} & \textbf{Spatial} & \textbf{Logical} & & \\
       % \cline{1-8}
       \midrule
      M$^{3}$CoT \cite{chen2024m} & 2024 & \checkmark & \checkmark & &  & Accuracy&11K\\
      MathVista \cite{lu2023mathvista} & 2024 &  & \checkmark & &  & Accuracy&6K\\
      MME-CoT \cite{jiang2025mme} & 2025 & \checkmark & \checkmark &\checkmark & \checkmark & Accuracy, Stablity, Efficiency&1.1K\\
      VISCO \cite{wu2024visco} & 2024& \checkmark & \checkmark &\checkmark & & Accuracy&1.6K\\
      MMIR \cite{yan2025multimodal} & 2025 &  &  & & \checkmark& Accuracy&0.5K\\
      EMMA \cite{hao2025can} & 2025&  & \checkmark & & & Accuracy&2.7K\\
      SpatialEval \cite{wang2024picture}& 2024&  &  & \checkmark& & Accuracy&4.6K\\
      MM-IQ \cite{cai2025mm}& 2025& \checkmark & \checkmark & \checkmark& \checkmark& Accuracy&2.7K\\
      ZeroBench \cite{roberts2025zerobench}& 2025&  & & \checkmark& & Accuracy&0.1K\\
      \bottomrule
      
    \end{tabular} }
  \end{center}
\end{table}

Multimodal reasoning benchmarks~\cite{cai2025mm,chen2024m,wu2024visco,jiang2025mme,lu2023mathvista,hessel2022abduction,hao2025can,yan2025multimodal} primarily focuses on four key task categories: general reasoning, academic-based reasoning, spatial reasoning, and logical reasoning. General reasoning tasks require models to utilize commonsense or general knowledge to answer questions effectively. Academic-based reasoning tasks involve subject-specific questions from areas such as science, mathematics and code, where models must apply appropriate principles and formulas. Spatial reasoning tasks evaluate a model’s ability to perceive and reason about spatial relationships between objects or elements. Logical reasoning tasks test a model’s capacity to manipulate logical symbols and understand semantic logic. The summary of benchmarks as shown in Table \ref{tab:tab2}.

\subsection{General Reasoning}
General reasoning tasks typically omit explicit information about target objects of the question, requiring models to rely on general knowledge or commonsense reasoning to produce correct answers. In \cite{cai2025mm}, the queries involve real-world entities such as vases, leaves, and animals. Multimodal models must classify these objects based on visual features, often drawing on external or implicit knowledge. To assess commonsense reasoning, \cite{chen2024m} refines and extends the Sherlock dataset \cite{hessel2022abduction} by generating questions, multiple-choice options, and corresponding answers grounded in visual cues, creating a benchmark specifically designed for commonsense reasoning evaluation.

\subsection{Academic-based Reasoning}
Academic reasoning tasks involve problems that require the application of subject-specific knowledge—such as mathematics, chemistry, and code—for multi-step reasoning. These tasks typically cannot be solved in a single step and are designed to assess a model’s ability to perform complex, structured reasoning~\cite{cai2025mm,chen2024m,wu2024visco,jiang2025mme,lu2023mathvista,hao2025can}.

To facilitate research in this area, several benchmarks have been proposed. For example, \cite{chen2024m} introduces a manually annotated dataset in science and mathematics, incorporating chain-of-thought (CoT) rationales. Visual-free samples are first filtered from existing datasets, and CoTs are manually annotated to strengthen the alignment between visual and textual reasoning. Similarly, \cite{cai2025mm} constructs a comprehensive benchmark using publicly available questions from the Chinese National Civil Service Exam.

Recent advances in large language models (LLMs) have enabled their use in preliminary data filtering, with subsequent human refinement to ensure high-quality annotations. \cite{lu2023mathvista} presents a benchmark covering diverse mathematical reasoning tasks—spanning algebra, geometry, logic, science, and statistics. Relevant data are selected using LLMs and then annotated by humans. In \cite{jiang2025mme}, a classification framework is proposed where LLMs compare model performance with and without CoT. Samples identified as requiring CoT are subsequently verified and annotated manually. \cite{hao2025can} contends that high-level cross-modal reasoning cannot be achieved through unimodal approaches alone. Accordingly, questions that can be answered solely based on image captions are removed. The remaining questions are categorized, and additional samples are curated to ensure a balanced distribution across task types.

Given its significant impact on reasoning, the ability to critique and correct CoT is essential. To address this, \cite{wu2024visco} introduces the VISCO dataset, which combines outputs from large vision-language models (LVLMs) with human verification. CoTs are generated and segmented into individual steps, each manually validated. Errors are annotated with explanations and corrected responses, enabling comprehensive assessment of reasoning refinement capabilities.

\subsection{Spatial Reasoning}
Spatial reasoning is a critical task for evaluating a model’s ability to understand and reason about spatial relationships. \cite{roberts2025zerobench} introduces a dataset comprising 100 complex spatial problems, aimed at assessing model performance in tasks such as counting and identifying points of intersection and localising points on maps. To increase task difficulty, a careful manual selection process was applied, ensuring the inclusion of visually complex elements and multi-step reasoning requirements. In addition, static relative spatial relationships—such as those found in Raven’s Progressive Matrices, visual analogies, inference, and grouping—are also central to spatial reasoning \cite{cai2025mm}.

These datasets typically adhere to the standard Visual Question Answering (VQA) paradigm, which involves pairs of images and questions. Extending these approaches, \cite{wang2024picture} proposed the Visual Text Question Answering (VTQA) framework, which incorporates textual descriptions of images alongside visual inputs. This enriched format covers a broader range of spatial reasoning skills, including relational understanding, navigation, and enumeration. Empirical results show that such additional textual information significantly improves model accuracy.

\subsection{Logical Reasoning}

Logical reasoning tasks can be broadly categorized into two main types: logical operations and semantic logic. Tasks involving logical operations primarily evaluate a model’s ability to apply formal logic operators such as AND, OR, and XOR. These tasks generally require the model to identify visual patterns encoding abstract logical operations, induce the underlying rules, and apply these rules to derive the correct output \cite{cai2025mm}. In contrast, semantic logic tasks focus on detecting cross-modal inconsistencies within complex, multi-element layouts. These inconsistencies manifest in five primary forms: factual contradictions, identity attribution errors, contextual mismatches, numerical discrepancies, and temporal/spatial inconsistencies. To develop a benchmark for such tasks, \cite{yan2025multimodal} employs GPT-o1 to generate semantically inconsistent elements across webpages, presentation slides, and posters. These synthetic inconsistencies are further refined using automated tools and subsequently verified by human annotators, culminating in a dataset of 534 meticulously curated and challenging samples.

\subsection{Metrics}
Comprehensive and diverse metrics are crucial for evaluating multimodal reasoning capabilities. Existing metrics include: Accuracy, Stability and Efficiency.
\begin{itemize}
    \item Accuracy: In multimodal reasoning tasks, it is important to evaluate not only the correctness of the final answer but also the accuracy of each thought in the reasoning process. Therefore, accuracy is defined as follows:
    \begin{equation}
        Accuracy=\frac{|T_{matched}|}{|T|}
    \end{equation}
where $T={t_1,t_2,\dots,t_N}$ denotes the COT, N denotes its length. $T_{matched}$ denotes the set of reasoning steps that match the groundtruth. Specifically, when directly measuring the answer, $N=1$. Accuracy is one of the most commonly used metrics and is widely adopted in multimodal reasoning research.
    \item Stablity: COT can enhance a model’s reasoning capabilities, but how it affects the model’s perception ability remains unknown. Stability measures the impact of CoT on the model’s perception ability: 
    \begin{equation}
        Stablity = Acc_{COT}^{P}- Acc_{DIR}^{P}
    \end{equation}
where $Acc_{COT}^{P}$ represents the accuracy on perception tasks with CoT, $Acc_{DIR}^{P}$ represents the accuracy when directly answering without CoT. 
    \item Efficiency : Reasoning Efficiency refers to the proportion of effective reasoning steps within the entire reasoning process. This metric was proposed by [4]. First, GPT-4o is used to evaluate whether each reasoning step contributes to generating the final answer. Then, the ratio of useful reasoning steps to all reasoning steps is calculated as follows:
\begin{equation}
    Efficiency = \frac{r-\alpha}{1-\alpha};
    r = \frac{|T_{relevant}|}{|T|}
\end{equation}

Where, ${T_relevant}$ represents the number of reasoning steps related to the answer, and ${\alpha}$ is a constant. Efficiency can directly reflect the information redundancy generated by the model and directly affect the generation speed of the model.

\end{itemize}

\section{Future Direction
}
\subsection{From Vision-Language Reasoning to Omni Reasoning}

Omni model is a unified framework that integrates multiple modalities, including text, image, audio, and video, to achieve comprehensive data processing and understanding. The impressive performance of GPT-4o \cite{gpt4o} demonstrates the significant potential of Omni models in perceiving and processing fully multimodal data, marking a notable achievement in \emph{Stage 1} (Chatbots, with conversational language). The subsequent introduction of R1-Omni \cite{zhao2025r1omniexplainableomnimultimodalemotion} indicates the transition from \emph{Stage 1} to \emph{Stage 2} (Reasoners, referring to human-level problem-solving ability) of Omni models. In contrast to vision-language (VL) models, Omni models extend to audio modality, which introduces substantial changes in both perception and state within Omni reasoning tasks.

\textbf{Perception: from Vision to Omni} \\ In contrasted with VL models, Omni models expand the reasoning space by incorporating diverse modal information, which consequently increases the complexity of reasoning tasks. During inference, Omni model is required to simultaneously perceive and integrate information from both vision and audio modalities to perform joint reasoning. There are two key challenges in Omni perception: 
1) Limited fine-grained cross-modal understanding: In audio and visual modalities, Omni models struggle with tasks that require deep contextual comprehension or involve subtle cross-modal nuances, leading to the use of incorrect or irrelevant information during reasoning. 
2) Inconsistent capability across modalities: The same task presented in different modalities can lead to inconsistent perceptions. Results from the Omni$\times$R benchmark \cite{chen2024omnixr} indicate that when a math problem is presented in text form, the model can generate correct answers with coherent reasoning. However, when the same problem is delivered via audio or video, the model may fail to produce reasoning altogether or generate incorrect answers. These modality-induced variations pose a significant challenge to Omni reasoning.

\textbf{State update: from VL generation to Omni generation} \\ Collaborative Multimodal Reasoning lies in simultaneously generating actions in the language space and updating states in the visual space. When extended to Omni models, state updates should consider extending beyond the visual domain to the audio space, capturing additional information essential for accurate reasoning. In tasks driven purely by audio inputs (e.g., ASR or speech translation), the model can generate reasoning outputs solely in the audio modality. However, when audio and visual modalities are jointly involved, both audio and visual state changes are necessary. For example, given the state "It is raining outside", an image may show only an indoor scene. The presence of background rain sounds in the audio stream can provide critical context for inferring external weather conditions, thus improving the quality and completeness of reasoning. 
Enabling such multimodal state updates in Omni models presents a significant challenge: the model must be capable of both understanding and generating data across all modalities, which goes beyond standard cross-modal alignment. Consequently, progress in Omni reasoning depends not only on advances in reasoning algorithms but also on improving the perceptual and generative capabilities of Omni model.

\subsection{From Multimodal Reasoning to Multimodal Agents}

OpenAI’s roadmap to AGI consists of five progressive stages, with the transition from \emph{Stage 2} to \emph{Stage 3} (Agents, where systems can take actions) marking a critical shift. Reasoners primarily focus on internal cognitive reasoning, meaning they engage in complex reasoning and problem-solving based on given information. In contrast, Agents possess the ability to take actions in an open environment, expanding from mere "thinking" to actual "execution". This transition from internal cognition to external action underpins the subsequent discussions on action generation and state updates. 

\textbf{Action generation: from Reasoning-driven to Interaction-driven} \\ In multimodal reasoning tasks, action generation does not directly interact with the environment but instead manifests as generating intermediate reasoning steps or conclusions based on multimodal inputs. Unlike reasoning models that operate within an internal cognitive framework, agents must execute actions and adapt to the resulting environmental changes. For example, in a web browsing task, an agent must generate executable actions (such as "click the login button") based on the web page's visual structure and text content while also adapting to changes in the page and making subsequent decisions accordingly. To achieve this transition, reinforcement learning plays a crucial role in bridging the gap between reasoning and autonomous decision-making. By continuously optimizing its actions based on environmental feedback, an agent can improve its decision-making process, ensuring that its interactions are not only active but also strategically planned for long-term goals \cite{deepresearch}.

\textbf{State update: from Cognition-driven to Environment-driven} \\ In multimodal reasoning, state updates primarily involve adjustments in cognitive representations, where the model dynamically refines its internal reasoning process to maintain logical coherence or enhance inferential depth. For instance, in computational tasks, adding auxiliary lines can help the model achieve a deeper understanding of a given problem. However, since agents’ actions directly alter the external environment, their state updates are determined by real-world changes, and often unpredictable. A key advancement in enabling effective state updates is environment modeling, which allows agents to transition from static reasoning to dynamic interaction. By building a structured representation of the external environment, such as a sandbox~\cite{zhou2023webarena,koh2024visualwebarena} or a world model~\cite{tu2025role,hao2023reasoning}, the agents are enabling more adaptive and proactive decision-making in open-ended scenarios \cite{operator}. 

\section{Conclusions}
This survey categorized multimodal reasoning into two levels based on the role of multimodal information in the reasoning process: \emph{Language-centric Multimodal Reasoning} and \emph{Collaborative Multimodal Reasoning}. This classification highlights the progression from pure language reasoning to comprehensive multimodal reasoning, offering a framework to analyze and anticipate future development directions. Specifically, we examined two promising directions -- \emph{Omni Reasoning} and \emph{Multimodal Agents} -- which focus on expanding the diversity of modalities in reasoning and merging reasoning with action, respectively.

However, we argue that advanced multimodal reasoning and multimodal agents, similar to language reasoning supported by strong language models and agent models relying on strong reasoning models \cite{zhang2025agentmodelsinternalizingchainofaction}, must be grounded in superior multimodal foundation models. Current Multimodal Large Language Models (MLLMs) are still fundamentally anchored to language priors. Substantive progress in multimodal reasoning requires unified multimodal understanding and generation, as well as continued exploration of joint multimodal pretraining paradigms that transcend language-centric biases. This shift will establish inherent cross-modal interactions rather than treating vision as an auxiliary signal to language reasoning, thereby enabling native and deeper multimodal reasoning capabilities that approach human-like cross-modal cognition.

\newpage
\bibliographystyle{plain} 
\bibliography{neurips_2024.bib}

\begin{thebibliography}{10}

\bibitem{bai2025qwen2}
Shuai Bai, Keqin Chen, Xuejing Liu, Jialin Wang, Wenbin Ge, Sibo Song, Kai Dang, Peng Wang, Shijie Wang, Jun Tang, et~al.
\newblock Qwen2. 5-vl technical report.
\newblock {\em arXiv preprint arXiv:2502.13923}, 2025.

\bibitem{bandyopadhyay2025thinking}
Dibyanayan Bandyopadhyay, Soham Bhattacharjee, and Asif Ekbal.
\newblock Thinking machines: A survey of llm based reasoning strategies.
\newblock {\em arXiv preprint arXiv:2503.10814}, 2025.

\bibitem{cai2025mm}
Huanqia Cai, Yijun Yang, and Winston Hu.
\newblock Mm-iq: Benchmarking human-like abstraction and reasoning in multimodal models.
\newblock {\em arXiv preprint arXiv:2502.00698}, 2025.

\bibitem{chen2024lion}
Gongwei Chen, Leyang Shen, Rui Shao, Xiang Deng, and Liqiang Nie.
\newblock Lion: Empowering multimodal large language model with dual-level visual knowledge.
\newblock In {\em Proceedings of the IEEE/CVF Conference on Computer Vision and Pattern Recognition}, pages 26540--26550, 2024.

\bibitem{chen2022visualgpt}
Jun Chen, Han Guo, Kai Yi, Boyang Li, and Mohamed Elhoseiny.
\newblock Visualgpt: Data-efficient adaptation of pretrained language models for image captioning.
\newblock In {\em Proceedings of the IEEE/CVF conference on computer vision and pattern recognition}, pages 18030--18040, 2022.

\bibitem{chen2025vinci}
Liang Chen, Lei Li, Haozhe Zhao, and Yifan Song.
\newblock R1-v: Reinforcing super generalization ability in vision-language models with less than 3, 2025.

\bibitem{chen2024omnixr}
Lichang Chen, Hexiang Hu, Mingda Zhang, Yiwen Chen, Zifeng Wang, Yandong Li, Pranav Shyam, Tianyi Zhou, Heng Huang, Ming-Hsuan Yang, et~al.
\newblock Omnixr: Evaluating omni-modality language models on reasoning across modalities.
\newblock {\em arXiv preprint arXiv:2410.12219}, 2024.

\bibitem{chen2025towards}
Qiguang Chen, Libo Qin, Jinhao Liu, Dengyun Peng, Jiannan Guan, Peng Wang, Mengkang Hu, Yuhang Zhou, Te~Gao, and Wangxiang Che.
\newblock Towards reasoning era: A survey of long chain-of-thought for reasoning large language models.
\newblock {\em arXiv preprint arXiv:2503.09567}, 2025.

\bibitem{chen2024m}
Qiguang Chen, Libo Qin, Jin Zhang, Zhi Chen, Xiao Xu, and Wanxiang Che.
\newblock M3cot: A novel benchmark for multi-domain multi-step multi-modal chain-of-thought.
\newblock {\em arXiv preprint arXiv:2405.16473}, 2024.

\bibitem{chen2024internvl}
Zhe Chen, Jiannan Wu, Wenhai Wang, Weijie Su, Guo Chen, Sen Xing, Muyan Zhong, Qinglong Zhang, Xizhou Zhu, Lewei Lu, et~al.
\newblock Internvl: Scaling up vision foundation models and aligning for generic visual-linguistic tasks.
\newblock In {\em Proceedings of the IEEE/CVF conference on computer vision and pattern recognition}, pages 24185--24198, 2024.

\bibitem{chen2023mitigating}
Zhiyang Chen, Yousong Zhu, Yufei Zhan, Zhaowen Li, Chaoyang Zhao, Jinqiao Wang, and Ming Tang.
\newblock Mitigating hallucination in visual language models with visual supervision.
\newblock {\em arXiv preprint arXiv:2311.16479}, 2023.

\bibitem{chen2024knowledge}
Zhuo Chen, Yichi Zhang, Yin Fang, Yuxia Geng, Lingbing Guo, Xiang Chen, Qian Li, Wen Zhang, Jiaoyan Chen, Yushan Zhu, et~al.
\newblock Knowledge graphs meet multi-modal learning: A comprehensive survey.
\newblock {\em arXiv preprint arXiv:2402.05391}, 2024.

\bibitem{cheng2024least}
Chuanqi Cheng, Jian Guan, Wei Wu, and Rui Yan.
\newblock From the least to the most: Building a plug-and-play visual reasoner via data synthesis.
\newblock {\em arXiv preprint arXiv:2406.19934}, 2024.

\bibitem{cheng2024vision}
Kanzhi Cheng, Yantao Li, Fangzhi Xu, Jianbing Zhang, Hao Zhou, and Yang Liu.
\newblock Vision-language models can self-improve reasoning via reflection.
\newblock {\em arXiv preprint arXiv:2411.00855}, 2024.

\bibitem{dong2024insight}
Yuhao Dong, Zuyan Liu, Hai-Long Sun, Jingkang Yang, Winston Hu, Yongming Rao, and Ziwei Liu.
\newblock Insight-v: Exploring long-chain visual reasoning with multimodal large language models.
\newblock {\em arXiv preprint arXiv:2411.14432}, 2024.

\bibitem{gao2024interleaved}
Jun Gao, Yongqi Li, Ziqiang Cao, and Wenjie Li.
\newblock Interleaved-modal chain-of-thought.
\newblock {\em arXiv preprint arXiv:2411.19488}, 2024.

\bibitem{gao2024cantor}
Timin Gao, Peixian Chen, Mengdan Zhang, Chaoyou Fu, Yunhang Shen, Yan Zhang, Shengchuan Zhang, Xiawu Zheng, Xing Sun, Liujuan Cao, et~al.
\newblock Cantor: Inspiring multimodal chain-of-thought of mllm.
\newblock In {\em Proceedings of the 32nd ACM International Conference on Multimedia}, pages 9096--9105, 2024.

\bibitem{guo2025deepseek}
Daya Guo, Dejian Yang, Haowei Zhang, Junxiao Song, Ruoyu Zhang, Runxin Xu, Qihao Zhu, Shirong Ma, Peiyi Wang, Xiao Bi, et~al.
\newblock Deepseek-r1: Incentivizing reasoning capability in llms via reinforcement learning.
\newblock {\em arXiv preprint arXiv:2501.12948}, 2025.

\bibitem{hao2023reasoning}
Shibo Hao, Yi~Gu, Haodi Ma, Joshua~Jiahua Hong, Zhen Wang, Daisy~Zhe Wang, and Zhiting Hu.
\newblock Reasoning with language model is planning with world model.
\newblock {\em arXiv preprint arXiv:2305.14992}, 2023.

\bibitem{hao2025can}
Yunzhuo Hao, Jiawei Gu, Huichen~Will Wang, Linjie Li, Zhengyuan Yang, Lijuan Wang, and Yu~Cheng.
\newblock Can mllms reason in multimodality? emma: An enhanced multimodal reasoning benchmark.
\newblock {\em arXiv preprint arXiv:2501.05444}, 2025.

\bibitem{hessel2022abduction}
Jack Hessel, Jena~D Hwang, Jae~Sung Park, Rowan Zellers, Chandra Bhagavatula, Anna Rohrbach, Kate Saenko, and Yejin Choi.
\newblock The abduction of sherlock holmes: A dataset for visual abductive reasoning.
\newblock In {\em European Conference on Computer Vision}, pages 558--575. Springer, 2022.

\bibitem{hu2024minicpm}
Shengding Hu, Yuge Tu, Xu~Han, Chaoqun He, Ganqu Cui, Xiang Long, Zhi Zheng, Yewei Fang, Yuxiang Huang, Weilin Zhao, et~al.
\newblock Minicpm: Unveiling the potential of small language models with scalable training strategies.
\newblock {\em arXiv preprint arXiv:2404.06395}, 2024.

\bibitem{hu2024visual}
Yushi Hu, Weijia Shi, Xingyu Fu, Dan Roth, Mari Ostendorf, Luke Zettlemoyer, Noah~A Smith, and Ranjay Krishna.
\newblock Visual sketchpad: Sketching as a visual chain of thought for multimodal language models.
\newblock {\em arXiv preprint arXiv:2406.09403}, 2024.

\bibitem{huang2024opera}
Qidong Huang, Xiaoyi Dong, Pan Zhang, Bin Wang, Conghui He, Jiaqi Wang, Dahua Lin, Weiming Zhang, and Nenghai Yu.
\newblock Opera: Alleviating hallucination in multi-modal large language models via over-trust penalty and retrospection-allocation.
\newblock In {\em Proceedings of the IEEE/CVF Conference on Computer Vision and Pattern Recognition}, pages 13418--13427, 2024.

\bibitem{huang2025vision}
Wenxuan Huang, Bohan Jia, Zijie Zhai, Shaosheng Cao, Zheyu Ye, Fei Zhao, Yao Hu, and Shaohui Lin.
\newblock Vision-r1: Incentivizing reasoning capability in multimodal large language models.
\newblock {\em arXiv preprint arXiv:2503.06749}, 2025.

\bibitem{jiang2025mme}
Dongzhi Jiang, Renrui Zhang, Ziyu Guo, Yanwei Li, Yu~Qi, Xinyan Chen, Liuhui Wang, Jianhan Jin, Claire Guo, Shen Yan, et~al.
\newblock Mme-cot: Benchmarking chain-of-thought in large multimodal models for reasoning quality, robustness, and efficiency.
\newblock {\em arXiv preprint arXiv:2502.09621}, 2025.

\bibitem{ke2024hydra}
Fucai Ke, Zhixi Cai, Simindokht Jahangard, Weiqing Wang, Pari~Delir Haghighi, and Hamid Rezatofighi.
\newblock Hydra: A hyper agent for dynamic compositional visual reasoning.
\newblock In {\em European Conference on Computer Vision}, pages 132--149. Springer, 2024.

\bibitem{koh2024visualwebarena}
Jing~Yu Koh, Robert Lo, Lawrence Jang, Vikram Duvvur, Ming~Chong Lim, Po-Yu Huang, Graham Neubig, Shuyan Zhou, Ruslan Salakhutdinov, and Daniel Fried.
\newblock Visualwebarena: Evaluating multimodal agents on realistic visual web tasks.
\newblock {\em arXiv preprint arXiv:2401.13649}, 2024.

\bibitem{lee2024multimodal}
Junlin Lee, Yequan Wang, Jing Li, and Min Zhang.
\newblock Multimodal reasoning with multimodal knowledge graph.
\newblock {\em arXiv preprint arXiv:2406.02030}, 2024.

\bibitem{li2024llava}
Bo~Li, Yuanhan Zhang, Dong Guo, Renrui Zhang, Feng Li, Hao Zhang, Kaichen Zhang, Peiyuan Zhang, Yanwei Li, Ziwei Liu, et~al.
\newblock Llava-onevision: Easy visual task transfer.
\newblock {\em arXiv preprint arXiv:2408.03326}, 2024.

\bibitem{li2025imagine}
Chengzu Li, Wenshan Wu, Huanyu Zhang, Yan Xia, Shaoguang Mao, Li~Dong, Ivan Vuli{\'c}, and Furu Wei.
\newblock Imagine while reasoning in space: Multimodal visualization-of-thought.
\newblock {\em arXiv preprint arXiv:2501.07542}, 2025.

\bibitem{li2024vocot}
Zejun Li, Ruipu Luo, Jiwen Zhang, Minghui Qiu, and Zhongyu Wei.
\newblock Vocot: Unleashing visually grounded multi-step reasoning in large multi-modal models.
\newblock {\em arXiv preprint arXiv:2405.16919}, 2024.

\bibitem{lin2025investigating}
Yujie Lin, Ante Wang, Moye Chen, Jingyao Liu, Hao Liu, Jinsong Su, and Xinyan Xiao.
\newblock Investigating inference-time scaling for chain of multi-modal thought: A preliminary study.
\newblock {\em arXiv preprint arXiv:2502.11514}, 2025.

\bibitem{liu2025visual}
Ziyu Liu, Zeyi Sun, Yuhang Zang, Xiaoyi Dong, Yuhang Cao, Haodong Duan, Dahua Lin, and Jiaqi Wang.
\newblock Visual-rft: Visual reinforcement fine-tuning.
\newblock {\em arXiv preprint arXiv:2503.01785}, 2025.

\bibitem{lu2023mathvista}
Pan Lu, Hritik Bansal, Tony Xia, Jiacheng Liu, Chunyuan Li, Hannaneh Hajishirzi, Hao Cheng, Kai-Wei Chang, Michel Galley, and Jianfeng Gao.
\newblock Mathvista: Evaluating mathematical reasoning of foundation models in visual contexts.
\newblock {\em arXiv preprint arXiv:2310.02255}, 2023.

\bibitem{mitra2024compositional}
Chancharik Mitra, Brandon Huang, Trevor Darrell, and Roei Herzig.
\newblock Compositional chain-of-thought prompting for large multimodal models.
\newblock In {\em Proceedings of the IEEE/CVF Conference on Computer Vision and Pattern Recognition}, pages 14420--14431, 2024.

\bibitem{ni2024visual}
Minheng Ni, Yutao Fan, Lei Zhang, and Wangmeng Zuo.
\newblock Visual-o1: Understanding ambiguous instructions via multi-modal multi-turn chain-of-thoughts reasoning.
\newblock {\em arXiv preprint arXiv:2410.03321}, 2024.

\bibitem{gpt4o}
Openai.
\newblock Hello gpt-4o.
\newblock {\em https://openai.com/index/hello-gpt-4o/}, 2023.

\bibitem{gpto1}
Openai.
\newblock Introducing openai o1-preview.
\newblock {\em https://openai.com/index/introducing-openai-o1-preview/}, 2023.

\bibitem{deepresearch}
Openai.
\newblock Introducing deep research.
\newblock {\em https://openai.com/index/introducing-deep-research/}, 2025.

\bibitem{operator}
Openai.
\newblock Introducing operator.
\newblock {\em https://openai.com/index/introducing-operator/}, 2025.

\bibitem{skywork2025r1v}
Yi~Peng.
\newblock Skywork r1v:pioneering multimodal reasoning with chain-of-thought.
\newblock {\em https://github.com/SkyworkAI/Skywork-R1V/}, 2025.

\bibitem{plaat2024reasoning}
Aske Plaat, Annie Wong, Suzan Verberne, Joost Broekens, Niki van Stein, and Thomas Back.
\newblock Reasoning with large language models, a survey.
\newblock {\em arXiv preprint arXiv:2407.11511}, 2024.

\bibitem{roberts2025zerobench}
Jonathan Roberts, Mohammad~Reza Taesiri, Ansh Sharma, Akash Gupta, Samuel Roberts, Ioana Croitoru, Simion-Vlad Bogolin, Jialu Tang, Florian Langer, Vyas Raina, et~al.
\newblock Zerobench: An impossible visual benchmark for contemporary large multimodal models.
\newblock {\em arXiv preprint arXiv:2502.09696}, 2025.

\bibitem{schrittwieser2020mastering}
Julian Schrittwieser, Ioannis Antonoglou, Thomas Hubert, Karen Simonyan, Laurent Sifre, Simon Schmitt, Arthur Guez, Edward Lockhart, Demis Hassabis, Thore Graepel, et~al.
\newblock Mastering atari, go, chess and shogi by planning with a learned model.
\newblock {\em Nature}, 588(7839):604--609, 2020.

\bibitem{shao2024visual}
Hao Shao, Shengju Qian, Han Xiao, Guanglu Song, Zhuofan Zong, Letian Wang, Yu~Liu, and Hongsheng Li.
\newblock Visual cot: Advancing multi-modal language models with a comprehensive dataset and benchmark for chain-of-thought reasoning.
\newblock {\em Advances in Neural Information Processing Systems}, 37:8612--8642, 2024.

\bibitem{sun2024visual}
Guangyan Sun, Mingyu Jin, Zhenting Wang, Cheng-Long Wang, Siqi Ma, Qifan Wang, Tong Geng, Ying~Nian Wu, Yongfeng Zhang, and Dongfang Liu.
\newblock Visual agents as fast and slow thinkers.
\newblock {\em arXiv preprint arXiv:2408.08862}, 2024.

\bibitem{team2025kimi}
Kimi Team, Angang Du, Bofei Gao, Bowei Xing, Changjiu Jiang, Cheng Chen, Cheng Li, Chenjun Xiao, Chenzhuang Du, Chonghua Liao, et~al.
\newblock Kimi k1. 5: Scaling reinforcement learning with llms.
\newblock {\em arXiv preprint arXiv:2501.12599}, 2025.

\bibitem{qvq-72b-preview}
Qwen Team.
\newblock Qvq: To see the world with wisdom, December 2024.

\bibitem{thawakar2025llamav}
Omkar Thawakar, Dinura Dissanayake, Ketan More, Ritesh Thawkar, Ahmed Heakl, Noor Ahsan, Yuhao Li, Mohammed Zumri, Jean Lahoud, Rao~Muhammad Anwer, et~al.
\newblock Llamav-o1: Rethinking step-by-step visual reasoning in llms.
\newblock {\em arXiv preprint arXiv:2501.06186}, 2025.

\bibitem{tu2025role}
Sifan Tu, Xin Zhou, Dingkang Liang, Xingyu Jiang, Yumeng Zhang, Xiaofan Li, and Xiang Bai.
\newblock The role of world models in shaping autonomous driving: A comprehensive survey.
\newblock {\em arXiv preprint arXiv:2502.10498}, 2025.

\bibitem{wang2024picture}
Jiayu Wang, Yifei Ming, Zhenmei Shi, Vibhav Vineet, Xin Wang, Sharon Li, and Neel Joshi.
\newblock Is a picture worth a thousand words? delving into spatial reasoning for vision language models.
\newblock {\em Advances in Neural Information Processing Systems}, 37:75392--75421, 2024.

\bibitem{wang2025tutorial}
Jun Wang.
\newblock A tutorial on llm reasoning: Relevant methods behind chatgpt o1.
\newblock {\em arXiv preprint arXiv:2502.10867}, 2025.

\bibitem{wang2024enhancing}
Weiyun Wang, Zhe Chen, Wenhai Wang, Yue Cao, Yangzhou Liu, Zhangwei Gao, Jinguo Zhu, Xizhou Zhu, Lewei Lu, Yu~Qiao, et~al.
\newblock Enhancing the reasoning ability of multimodal large language models via mixed preference optimization.
\newblock {\em arXiv preprint arXiv:2411.10442}, 2024.

\bibitem{wu2025boosting}
Jinyang Wu, Mingkuan Feng, Shuai Zhang, Ruihan Jin, Feihu Che, Zengqi Wen, and Jianhua Tao.
\newblock Boosting multimodal reasoning with mcts-automated structured thinking.
\newblock {\em arXiv preprint arXiv:2502.02339}, 2025.

\bibitem{wu2024visco}
Xueqing Wu, Yuheng Ding, Bingxuan Li, Pan Lu, Da~Yin, Kai-Wei Chang, and Nanyun Peng.
\newblock Visco: Benchmarking fine-grained critique and correction towards self-improvement in visual reasoning.
\newblock {\em arXiv preprint arXiv:2412.02172}, 2024.

\bibitem{xiao2024enhancing}
Ziyang Xiao, Dongxiang Zhang, Xiongwei Han, Xiaojin Fu, Wing~Yin Yu, Tao Zhong, Sai Wu, Yuan Wang, Jianwei Yin, and Gang Chen.
\newblock Enhancing llm reasoning via vision-augmented prompting.
\newblock {\em Advances in Neural Information Processing Systems}, 37:28772--28797, 2024.

\bibitem{xu2025towards}
Fengli Xu, Qianyue Hao, Zefang Zong, Jingwei Wang, Yunke Zhang, Jingyi Wang, Xiaochong Lan, Jiahui Gong, Tianjian Ouyang, Fanjin Meng, et~al.
\newblock Towards large reasoning models: A survey of reinforced reasoning with large language models.
\newblock {\em arXiv preprint arXiv:2501.09686}, 2025.

\bibitem{xu2411llava}
Guowei Xu, Peng Jin, Li~Hao, Yibing Song, Lichao Sun, and Li~Yuan.
\newblock Llava-cot: Let vision language models reason step-by-step, 2024.
\newblock {\em arXiv preprint arXiv:2411.10440}, 2024.

\bibitem{yan2025multimodal}
Qianqi Yan, Yue Fan, Hongquan Li, Shan Jiang, Yang Zhao, Xinze Guan, Ching-Chen Kuo, and Xin~Eric Wang.
\newblock Multimodal inconsistency reasoning (mmir): A new benchmark for multimodal reasoning models.
\newblock {\em arXiv preprint arXiv:2502.16033}, 2025.

\bibitem{yang2025r1onevision}
Yi~Yang.
\newblock R1-onevision: Open-source multimodal large language model with reasoning ability, 2025.

\bibitem{yao2024mulberry}
Huanjin Yao, Jiaxing Huang, Wenhao Wu, Jingyi Zhang, Yibo Wang, Shunyu Liu, Yingjie Wang, Yuxin Song, Haocheng Feng, Li~Shen, et~al.
\newblock Mulberry: Empowering mllm with o1-like reasoning and reflection via collective monte carlo tree search.
\newblock {\em arXiv preprint arXiv:2412.18319}, 2024.

\bibitem{yin2023lamm}
Zhenfei Yin, Jiong Wang, Jianjian Cao, Zhelun Shi, Dingning Liu, Mukai Li, Xiaoshui Huang, Zhiyong Wang, Lu~Sheng, Lei Bai, et~al.
\newblock Lamm: Language-assisted multi-modal instruction-tuning dataset, framework, and benchmark.
\newblock {\em Advances in Neural Information Processing Systems}, 36:26650--26685, 2023.

\bibitem{zhang2022multimodal}
Ningyu Zhang, Lei Li, Xiang Chen, Xiaozhuan Liang, Shumin Deng, and Huajun Chen.
\newblock Multimodal analogical reasoning over knowledge graphs.
\newblock {\em arXiv preprint arXiv:2210.00312}, 2022.

\bibitem{zhang2024improve}
Ruohong Zhang, Bowen Zhang, Yanghao Li, Haotian Zhang, Zhiqing Sun, Zhe Gan, Yinfei Yang, Ruoming Pang, and Yiming Yang.
\newblock Improve vision language model chain-of-thought reasoning.
\newblock {\em arXiv preprint arXiv:2410.16198}, 2024.

\bibitem{zhang2024llm}
Yadong Zhang, Shaoguang Mao, Tao Ge, Xun Wang, Adrian de~Wynter, Yan Xia, Wenshan Wu, Ting Song, Man Lan, and Furu Wei.
\newblock Llm as a mastermind: A survey of strategic reasoning with large language models.
\newblock {\em arXiv preprint arXiv:2404.01230}, 2024.

\bibitem{zhang2025agentmodelsinternalizingchainofaction}
Yuxiang Zhang, Yuqi Yang, Jiangming Shu, Xinyan Wen, and Jitao Sang.
\newblock Agent models: Internalizing chain-of-action generation into reasoning models, 2025.

\bibitem{zhang2023multimodal}
Zhuosheng Zhang, Aston Zhang, Mu~Li, Hai Zhao, George Karypis, and Alex Smola.
\newblock Multimodal chain-of-thought reasoning in language models.
\newblock {\em arXiv preprint arXiv:2302.00923}, 2023.

\bibitem{zhao2025r1omniexplainableomnimultimodalemotion}
Jiaxing Zhao, Xihan Wei, and Liefeng Bo.
\newblock R1-omni: Explainable omni-multimodal emotion recognition with reinforcement learning.
\newblock {\em arXiv preprint arXiv:2503.05379}, 2025.

\bibitem{zhao2024mg}
Xiangyu Zhao, Xiangtai Li, Haodong Duan, Haian Huang, Yining Li, Kai Chen, and Hua Yang.
\newblock Mg-llava: Towards multi-granularity visual instruction tuning.
\newblock {\em arXiv preprint arXiv:2406.17770}, 2024.

\bibitem{zheng2023ddcot}
Ge~Zheng, Bin Yang, Jiajin Tang, Hong-Yu Zhou, and Sibei Yang.
\newblock Ddcot: Duty-distinct chain-of-thought prompting for multimodal reasoning in language models.
\newblock {\em Advances in Neural Information Processing Systems}, 36:5168--5191, 2023.

\bibitem{zhou2025r1}
Hengguang Zhou, Xirui Li, Ruochen Wang, Minhao Cheng, Tianyi Zhou, and Cho-Jui Hsieh.
\newblock R1-zero's" aha moment" in visual reasoning on a 2b non-sft model.
\newblock {\em arXiv preprint arXiv:2503.05132}, 2025.

\bibitem{zhou2024image}
Qiji Zhou, Ruochen Zhou, Zike Hu, Panzhong Lu, Siyang Gao, and Yue Zhang.
\newblock Image-of-thought prompting for visual reasoning refinement in multimodal large language models.
\newblock {\em arXiv preprint arXiv:2405.13872}, 2024.

\bibitem{zhou2023webarena}
Shuyan Zhou, Frank~F Xu, Hao Zhu, Xuhui Zhou, Robert Lo, Abishek Sridhar, Xianyi Cheng, Tianyue Ou, Yonatan Bisk, Daniel Fried, et~al.
\newblock Webarena: A realistic web environment for building autonomous agents.
\newblock {\em arXiv preprint arXiv:2307.13854}, 2023.

\bibitem{zhou2025they}
Yikang Zhou, Tao Zhang, Shilin Xu, Shihao Chen, Qianyu Zhou, Yunhai Tong, Shunping Ji, Jiangning Zhang, Xiangtai Li, and Lu~Qi.
\newblock Are they the same? exploring visual correspondence shortcomings of multimodal llms.
\newblock {\em arXiv preprint arXiv:2501.04670}, 2025.

\bibitem{zhu2022multimodal}
Chaoyu Zhu, Zhihao Yang, Xiaoqiong Xia, Nan Li, Fan Zhong, and Lei Liu.
\newblock Multimodal reasoning based on knowledge graph embedding for specific diseases.
\newblock {\em Bioinformatics}, 38(8):2235--2245, 2022.

\bibitem{zhu2022multi}
Xiangru Zhu, Zhixu Li, Xiaodan Wang, Xueyao Jiang, Penglei Sun, Xuwu Wang, Yanghua Xiao, and Nicholas~Jing Yuan.
\newblock Multi-modal knowledge graph construction and application: A survey.
\newblock {\em IEEE Transactions on Knowledge and Data Engineering}, 36(2):715--735, 2022.

\end{thebibliography}

\end{document}